\documentclass[letterpaper, 10 pt, conference]{ieeeconf}

\IEEEoverridecommandlockouts
\usepackage{times}

\usepackage{multicol}

\usepackage{balance}
\usepackage{xparse} 
\usepackage{amsmath}
\usepackage{amssymb}
\usepackage{amsfonts} 
\usepackage{dsfont}
\usepackage{graphicx}
\usepackage{url}
\usepackage{booktabs}
\usepackage[para]{threeparttable}
\usepackage{multirow}
\usepackage{makecell}
\usepackage{cite}

\usepackage[colorlinks,bookmarksopen,bookmarksnumbered,citecolor=red,urlcolor=red]{hyperref}
\usepackage{cleveref}
\usepackage{xcolor}
\usepackage{framed}
\usepackage{caption}
\usepackage{subcaption}

\usepackage{algorithm}
\usepackage[noend]{algpseudocode}

\algrenewcommand\algorithmicrequire{\textbf{Input:}}
\algrenewcommand\algorithmicensure{\textbf{Output:}}
\algnewcommand\algorithmicinput{\textbf{Input:}}
\algnewcommand\INPUT{\item[\algorithmicinput]}
\usepackage{setspace}

\colorlet{shadecolor}{gray!10}

\usepackage{xparse}

\NewDocumentCommand\bbm{}{ \begin{bmatrix} }
\NewDocumentCommand\ebm{}{ \end{bmatrix} }
\NewDocumentCommand\Vector{m}{ \boldsymbol{\mathbf{#1}} }

\NewDocumentCommand\Matrix{m}{ \boldsymbol{\mathbf{#1}} }

\NewDocumentCommand\Real{}{ \mathbb{R} }

\NewDocumentCommand\LieGroupO{m}{ \mathrm{O}(#1) }

\NewDocumentCommand\T{}{\mathsf{T}}

\NewDocumentCommand\diag{m}{\text{diag}\left(#1\right)}

\title{Inverse Kinematics as Low-Rank Euclidean Distance Matrix Completion}
\author{Filip Mari\'{c}$^{1,2}$, Matthew Giamou$^{1}$, Ivan Petrovi\'{c}$^{2}$, and Jonathan Kelly$^{1}$%
\thanks{$^{1}$Authors are with the Space \& Terrestrial Autonomous Robotic Systems (STARS) Laboratory at the University of Toronto Institute for Aerospace Studies (UTIAS), Toronto, Ontario, Canada, M3H~5T6. Email: \texttt{<first name>.<last name>@robotics.utias.utoronto.ca}}
\thanks{$^{2}$ F. Mari\'{c} and I. Petrovi\'{c} are with the Laboratory for Autonomous Systems and Mobile Robotics (LAMOR) at the University of Zagreb, Croatia.}
}

\begin{document}
\maketitle

\begin{abstract}
The majority of inverse kinematics (IK) algorithms search for solutions in a configuration space defined by joint angles.
 However, the kinematics of many robots can also be described in terms of distances between rigidly-attached points, which collectively form a Euclidean distance matrix.
 This alternative geometric description of the kinematics reveals an elegant equivalence between IK and the problem of low-rank matrix completion.
 We use this connection to implement a novel Riemannian optimization-based solution to IK for various articulated robots with symmetric joint angle constraints.
\end{abstract}
\section{Introduction}
Inverse kinematics (IK) algorithms play an essential role in robot motion and task planning.
For robots with redundant degrees of freedom (DOFs), such as manipulators, there exist infinitely many IK solutions and numerical methods are needed to find feasible configurations.
The vast majority of IK algorithms use a joint angle parameterization, with decision variables $\Vector{\Theta} \in \Real^n$ comprising the configuration space $\mathcal{C}$.
These variables are related to the workspace by
\begin{equation} \label{eq:kinematics}
	F(\Vector{\Theta}) = \Vector{w} \in \mathcal{T},
\end{equation}
where $F$ is the forward kinematics mapping and $\mathcal{T}$ is the task space containing the target poses $\Vector{w}$ of the end-effector.
For complex robots with many degrees of freedom, $F$ is a highly nonlinear function composed of trigonometric products and sums.
Thus, formulating IK as an optimization problem over decision variable $\Vector{\Theta}$ incorporates either a nonconvex cost function or nonconvex constraints.
These nonconvex formulations introduce local minima that complicate the search for a global minimum $\Vector{\Theta}^\star$ that exactly satisfies \Cref{eq:kinematics}.

\begin{figure}
  \includegraphics[width=\columnwidth]{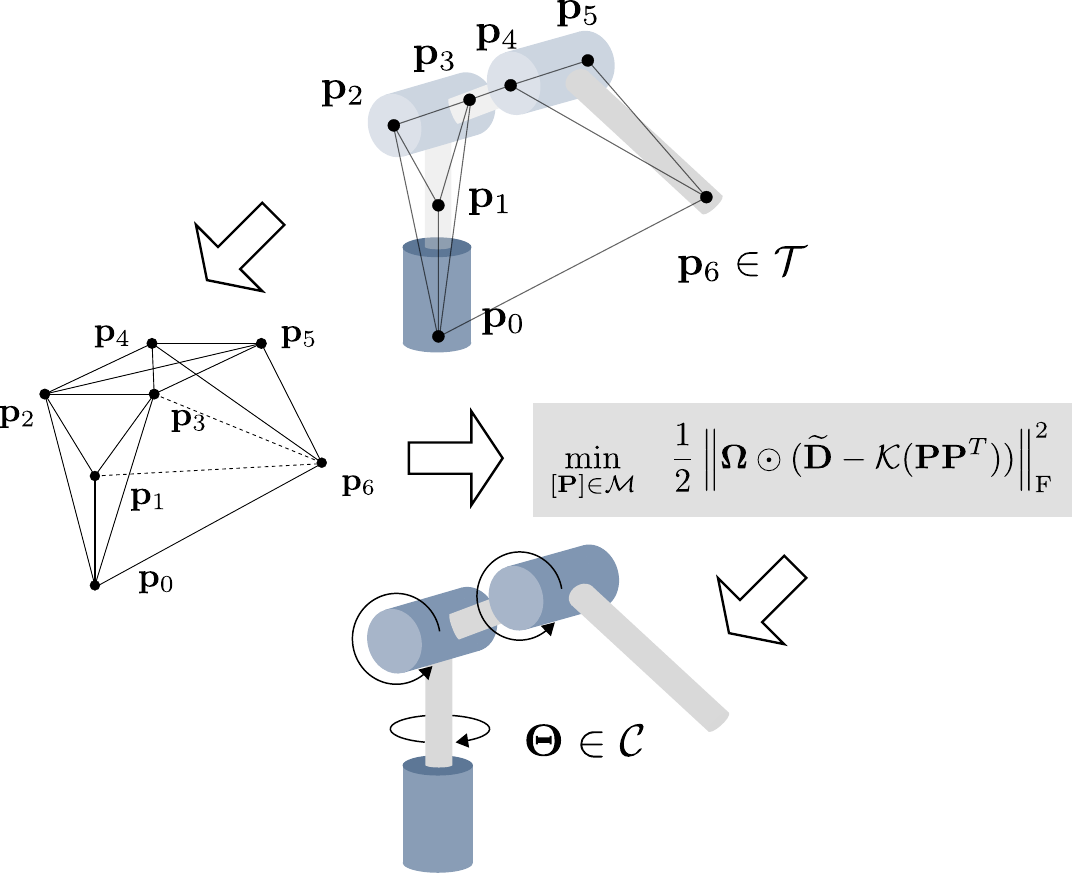}
  \vspace*{-3mm}
  \caption{The IK problem consists of finding the joint angles $\Vector{\Theta}$ that bring the end-effector of a 3-DOF robotic manipulator to its goal position ($\Vector{p}_{6}$ in the diagram). Our method uses distance information common to all possible IK solutions by defining an incomplete graph whose edges are weighted by known distances. Then, we use low-rank matrix completion to recover the weights corresponding to the unknown edges, solving the IK problem.}\label{fig:system_overview}
\vspace{-7mm}
\end{figure}

Recently, an alternative strategy for describing IK has emerged based on constraining the distances between points fixed to the links of articulated robots~\cite{blanchini_convex_2017,maricInverseKinematicsSerial2020,Le_Naour_2019}. 
Most notably, Porta et al.\ have cast inverse kinematics as a special case of the general Euclidean distance matrix (EDM) completion problem for several articulated robots~\cite{Porta2005, porta_branch-and-prune_2005}.
In this work, we apply the Riemannian optimization-based approach to low-rank EDM completion introduced in \cite{Mishra_2011} to the IK problem with symmetric joint limit constraints.
\Cref{fig:system_overview} illustrates our approach for a simple 3-DOF manipulator.  

\section{Formulation}
A collection of $N$ points rigidly attached to a manipulator with a $K$-dimensional workspace is arranged in a matrix ${\mathbf{P} = [\mathbf{p}_{0}, \mathbf{p}_{1},\dots ,\mathbf{p}_{N-1}]^\T \in \mathbb{R}^{N \times K}}$. 
All inter-point distances $d_{u,v}$ between $\Vector{p}_u$ and $\Vector{p}_v$ are determined via the Euclidean norm:
\begin{equation*}
	d_{u,v}= \|\mathbf{p}_{u} - \mathbf{p}_{v}\|_{2}\,.
\end{equation*}
The inner product $\mathbf{X} := \Matrix{P}\Matrix{P}^\T$ is a symmetric positive semidefinite \textit{Gram matrix} that can be used to compute the full set of squared inter-point distances $d_{u,v}^2$,
\begin{equation}\label{eq:EDM}
	\mathbf{D} = \mathcal{K}(\Matrix{X}) = 2 \ \text{Sym}(\diag{\Matrix{X}}\Matrix{1}^{\T} - \Matrix{X}),
\end{equation}
where $\text{Sym}(\Matrix{X}) = \frac{1}{2}(\Matrix{X} + \Matrix{X}^{\T})$ and $\diag{\Matrix{X}}$ is the vector formed by the main diagonal of $\Matrix{X}$.
The resulting matrix $\mathbf{D}$ is known as a \textit{Euclidean distance matrix} or EDM~\cite{dokmanic_euclidean_2015}.
Given a full EDM, we can use eigenvalue decomposition to extract the underlying point set $\Matrix{P}$ up to a rigid transformation.

In our point-based kinematics model, a subset of inter-point distances is defined by the manipulator's rigid link geometry and desired end-effector pose.
Joint angle limits define lower bounds on distances between points that share a neighbouring joint.
The problem of finding the remaining unknown entries in the resulting partial EDM is known as the \textit{EDM completion problem}~\cite{dokmanic_euclidean_2015, libertiEuclideanDistanceGeometry2014}.
We denote the $N\times N$ binary matrices that index the locations of all known and lower bounded squared distances of the partial EDM $\tilde{\Matrix{D}}$ as $\Omega$ and $\Psi$ respectively.
The problem of finding the missing EDM entries can now be expressed as low-rank matrix completion:
\begin{equation}\label{eq:EDM_IK}
  \begin{aligned}
  \min_{[\Matrix{P}]\in \mathcal{M}} \quad &\frac{1}{2}\left\lVert \Omega\odot(\widetilde{\Matrix{D}}-\mathcal{K}(\Matrix{P}\Matrix{P}^{T}))\right\rVert_{F}^{2}\\ + &\frac{1}{2}\left\lVert \mathrm{max}\left\{\Psi\odot(\widetilde{\Matrix{D}} - \mathcal{K}(\Matrix{P}\Matrix{P}^{T})), 0 \right\}\right\rVert_{F}^{2}\, ,
  \end{aligned}
\end{equation}
where $\odot$ is the element-wise Hadamard product, $\mathrm{max}$ is taken element-wise with 0, and $\mathcal{M}$ is the Riemannian manifold described in \Cref{eq:equivalence}.
By factoring the Gram matrix as $\Matrix{X}=\Matrix{P}\Matrix{P}^{\T}$ and solving for $\Matrix{P}$, we are implicitly constraining the dimensionality of the search space \cite{Burer_2004}, without changing the global minimum \cite{fangEuclideanDistanceMatrix2012}.

The factorization of the Gram matrix is invariant to orthogonal transformations of the point set $\Matrix{P} \mapsto \Matrix{P}\Matrix{Q}$, where $\Matrix{Q} \in \LieGroupO{K}$.
It follows that the cost function minima are not isolated, which significantly hinders the convergence of second-order optimization methods \cite{absil2009optimization}.
Following \cite{Mishra_2011}, the local search is constrained to the quotient manifold $\mathcal{M}\equiv\mathbb{R}_{*}^{N \times K}/ \LieGroupO{K}$, which represents the equivalence class
\begin{equation}\label{eq:equivalence}
	[\Matrix{P}] = \left\{\Matrix{P}\Matrix{Q} \vert \,\Matrix{Q} \in \mathbb{R}^{K \times K}\, , \Matrix{Q}^{T}\Matrix{Q}= \Matrix{I}\right\}.
\end{equation}
The geometry of the Riemannian manifold $\mathcal{M}$ isolates minima, and a variety of algorithms can be applied to solve the problem in \Cref{eq:EDM_IK} as an unconstrained Riemannian optimization problem \cite{absil2009optimization}.

\section{Results and Future Work}
\begin{table}
\renewcommand{\arraystretch}{1.0}
  \centering
  \caption{Experimental results. Each row summarizes 1000 trials. Asterisks ($\ast$) denote the use of angular joint limits.}\label{table}
  	\vspace{-1mm}
	\begin{tabular}{lcc}
    Robot    & Success (\%)  & Mean Runtime (ms)\\
    \hline
    \texttt{planar-10}      & 100    & 8.1   \\
    \texttt{planar-10}*      & 100    & 101.9   \\
    \texttt{planar-100}       & 100     & 166.5  \\
    \texttt{planar-100}*       & 100     & 1882  \\
    \texttt{spherical-10}      & 100    & 24.1   \\
    \texttt{spherical-10}*      & 100    & 58.4   \\
    \texttt{spherical-100}      & 100   & 641.7   \\
    \texttt{spherical-100}*      & 99.9    & 9759   \\
    \texttt{UR10}       & 84.2     & 462.5  \\
    \texttt{UR10}*       & 63.0     & 350.5  \\
  \end{tabular}
  \vspace{-6mm}
\end{table}
In \Cref{table}, we present the results of a simulated evaluation of our method on IK problems for a variety of manipulators.
In every instance we formulate and solve the problem in \Cref{eq:EDM_IK} in Python using the \texttt{pymanopt}\footnote{\url{github.com/pymanopt/pymanopt}} package on a laptop computer with an Intel\textsuperscript{\tiny \textregistered}	 Core\textsuperscript{\tiny{TM}} i7-8750H CPU at 2.20 GHz and 16 GB RAM.
The IK problems are generated by randomly sampling the configuration space within joint limits, and the problem is initialized using a set of points corresponding to a configuration with joint angles set to $0$.
Joint limits are random and symmetric about $0$, with a minimum value of $0.2$ radians.
A solution is considered to be correct (successful) if it achieves position and orientation errors lower than $0.01$ meters and $0.01$ radians, respectively.

We begin with 10 and 100 joint manipulators with unit link lengths, denoted by \texttt{planar-10}, \texttt{spherical-10}, \texttt{planar-100} and \texttt{spherical-100} in \Cref{table}.
In all experiments, our method achieves a $100 \%$ success rate with mean computation times well under a second.
Surprisingly, this result holds even in the somewhat extreme 100-joint examples, albeit with computation time an order of magnitude greater.
Next, we introduce joint limits and repeat the experiment, denoting the results with an asterisk ($\ast$).
Again, in all cases there is a $100 \%$ success rate, with the exception of one failure occurring for \texttt{spherical-100}*.
The computation times in the joint-limited cases are generally higher due to the introduction of the second cost term in \Cref{eq:EDM_IK} that enforces lower bounds on a subset of distances.
Finally, we take the practical example of the UR10 6-DOF manipulator, where our method succeeds in $84.2 \%$ and $63.0 \%$ of instances (without and with joint limits, respectively).
These success rates are comparable to solvers available in the Kinematics and Dynamics Library~\cite[Table 1]{beeson2015trac}.

These results show promise for distance geometric models in kinematics and motion planning, since low-rank matrix completion and the distance geometry problem in general \cite{libertiEuclideanDistanceGeometry2014} have strong theoretical underpinnings.
In our work, we seek to exploit this perspective to develop new methods for solving and analyzing complex inverse kinematics problems.
We are currently experimenting with spherical obstacles, which are easily incorporated into our formulation as points with fixed distances from the robot base and one another, and distances to points on the robot that are bounded from below by each obstacle's radius.

{\footnotesize
\bibliographystyle{ieeetr}
\bibliography{references}

\begin{thebibliography}{10}

\bibitem{blanchini_convex_2017}
F.~Blanchini, G.~Fenu, G.~Giordano, and F.~A. Pellegrino, ``A convex
  programming approach to the inverse kinematics problem for manipulators under
  constraints,'' {\em European Journal of Control}, vol.~33, pp.~11--23, Jan.
  2017.

\bibitem{maricInverseKinematicsSerial2020}
F.~Mari\'{c}, M.~Giamou, S.~Khoubyarian, I.~Petrovi\'{c}, and J.~Kelly,
  ``Inverse kinematics for serial kinematic chains via sum of squares
  optimization,'' in {\em {IEEE} International Conference on Robotics and
  Automation}, pp.~7101--7107, May/Jun. 2020.

\bibitem{Le_Naour_2019}
T.~Le~Naour, N.~Courty, and S.~Gibet, ``Kinematics in the metric space,'' {\em
  Computers \& Graphics}, vol.~84, p.~13–23, Nov 2019.

\bibitem{Porta2005}
J.~M. Porta, L.~Ros, and F.~Thomas, ``Inverse kinematics by distance matrix
  completion,'' in {\em Proceedings of the12th International Workshop on
  Computational Kinematics}, May 2005.

\bibitem{porta_branch-and-prune_2005}
J.~Porta, L.~Ros, F.~Thomas, and C.~Torras, ``A branch-and-prune solver for
  distance constraints,'' {\em IEEE Transactions on Robotics}, vol.~21,
  pp.~176--187, Apr. 2005.

\bibitem{Mishra_2011}
B.~Mishra, G.~Meyer, and R.~Sepulchre, ``Low-rank optimization for distance
  matrix completion,'' {\em IEEE Conference on Decision and Control and
  European Control Conference}, pp.~4455--4460, Dec 2011.

\bibitem{dokmanic_euclidean_2015}
I.~Dokmanic, R.~Parhizkar, J.~Ranieri, and M.~Vetterli, ``Euclidean {{Distance
  Matrices}}: {{Essential Theory}}, {{Algorithms}} and {{Applications}},'' {\em
  IEEE Signal Processing Magazine}, vol.~32, pp.~12--30, Nov. 2015.

\bibitem{libertiEuclideanDistanceGeometry2014}
L.~Liberti, C.~Lavor, N.~Maculan, and A.~Mucherino, ``Euclidean distance
  geometry and applications,'' {\em SIAM Rev.}, vol.~56, pp.~3--69, Jan. 2014.

\bibitem{Burer_2004}
S.~Burer and R.~D. Monteiro, ``Local minima and convergence in low-rank
  semidefinite programming,'' {\em Mathematical Programming}, vol.~103,
  p.~427–444, Dec 2004.

\bibitem{fangEuclideanDistanceMatrix2012}
H.~Fang and D.~P. O'Leary, ``Euclidean distance matrix completion problems,''
  {\em Optimization Methods and Software}, vol.~27, pp.~695--717, Oct. 2012.

\bibitem{absil2009optimization}
P.-A. Absil, R.~Mahony, and R.~Sepulchre, {\em Optimization Algorithms on
  Matrix Manifolds}.
\newblock Princeton University Press, 2009.

\bibitem{beeson2015trac}
P.~Beeson and B.~Ames, ``{TRAC-IK}: An open-source library for improved solving
  of generic inverse kinematics,'' in {\em IEEE-RAS International Conference on
  Humanoid Robots}, pp.~928--935, IEEE, 2015.

\end{thebibliography}
}

\end{document}